\documentclass{article}

\usepackage[preprint]{neurips_2024}

\usepackage[utf8]{inputenc} 
\usepackage[T1]{fontenc}    
\usepackage{hyperref}       
\usepackage{url}            
\usepackage{booktabs}       
\usepackage{amsfonts}       
\usepackage{nicefrac}       
\usepackage{microtype}      
\usepackage{xcolor}         

\usepackage{graphicx}
\usepackage{float} 
\usepackage{subfigure}
\usepackage{amsmath}
\usepackage{bbm}  
\usepackage{amssymb}
\usepackage{bm}
\usepackage{algorithm}
\usepackage{algpseudocode}
\usepackage{multirow}

\usepackage{appendix}
\usepackage{colortbl}

\usepackage{natbib}
\setcitestyle{numbers,square}

\usepackage{amsthm}  
\newtheorem{theorem}{Theorem}

\newtheorem{lemma}[theorem]{Lemma}

\newtheorem{definition}{Definition}[section]

\theoremstyle{remark}

\newcommand{\mysplit}[1]{%
  \begin{tabular}{@{}c@{}}   
    #1
  \end{tabular}
  }

\title{LFFR: Logistic Function For (single-output) Regression}

\author{%
\href{https://orcid.org/0000-0003-0378-0607}{\includegraphics[scale=0.06]{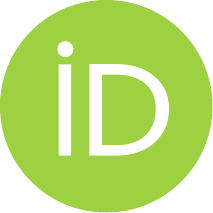}\hspace{1mm}John Chiang }
  \\
  \texttt{john.chiang.smith@gmail.com} \\
}

\begin{document}
\maketitle
\begin{abstract}
Privacy-preserving regression in machine learning is a crucial area of research, aimed at enabling the use of powerful machine learning techniques while protecting individuals' privacy. In this paper, we implement privacy-preserving regression training using data encrypted under a fully homomorphic encryption scheme. We first examine the common linear regression algorithm and propose a (simplified) fixed Hessian for linear regression training, which can be applied for any datasets even not normialized into the range $[0, 1]$. We also generalize this constant Hessian matrix to the ridge regression version, namely linear regression that includes a regularization term to penalize large coefficients.  However, our main contribution is to develop a novel and efficient algorithm called LFFR for homomorphic regression using the logistic function, which could model more complex relatiions between input values and the output prediction in compariosn with the linear regression. We also find a constant simplified Hessian to train our algorihtm LFFR using Newton-like method and compare it against to with our new fixed-Hessian linear regression taining over two real-world datasets. 
We suggest normalizing not only the data but also the target predictions even for the original linear regression used in a privacy-preserving manner, which is helpful to remain weights in small range, say $[-5, +5]$ good for refreshing ciphertext setting parameters, and avoid tuning the regualrizaion parameter $\lambda$ via cross validation. The linear regression with normalized predictions could be a viable alternative to ridge regression.
\end{abstract}

\section{Introduction}

\subsection{Background}
Regression algorithms have wide applications in various fields due to their ability to model and predict continuous outcomes. In a recommendation system, for example, the preferences and ratings of many users for various items are collected and subjected to a regression algorithm. This algorithm builds a model from a user's past ratings given on certain items as well as similar decisions made by other users. The model can then be used to predict how that user will rate other items of interest. A large number of algorithms exist for producing such predictive models, many of which are actively employed by large platforms such as Amazon and Netflix. Such algorithms are also used over large medical databases, financial data, etc.

Regression algorithms require access to users' private data in order to construct predictive models. This would raise significant concerns regarding data privacy breaches, thereby discouraging users from sharing their sensitive information. To address these challenges, several solutions have been proposed, among which homomorphic encryption stands out as one of the most secure approaches. This method allows computations to be performed on encrypted data without the need to decrypt it first, thereby preserving the privacy and confidentiality of the users' data.

There are various algorithms in machine learning for regression tasks. For example, linear regression and ridge regression are two commonly adopted choices. Given some sample points in high-dimensional space, linear regression or ridge regression outputs the best-fit linear curve through these points. However, for more complex problems, where the relationship between the independent and dependent variables is not linear, these two algorithms may perform badly. In this paper we present a new efficient algorithm termed LFFR based on a logistic regression-like simple 2-layer neural networks to deal with the non-linear cases. Our goal is to develop some novel algorithms that can leadge the (Simplified) Fixed Hessian~\cite{IDASH2018bonte} methods to be employed in the encrypted state. 

\subsection{Related work}
Unlike privacy-preserving neural network inference~\cite{,, gilad2016cryptonets, li2018privacy, bourse2018fast, chiang2022volleyrevolver, chillotti2021programmable,kim2018matrix,chabanne2017privacy} and homomorphic logistic regression training~\cite{chiang2022privacy,chiang2024privacy, kim2018logistic,kim2018secure,IDASH2018bonte,aono2016scalable,IDASH2018gentry,IDASH2019kim}, which have seen extensive research and development, including contributions from the iDASH competitions in 2018 and 2019, regression learning without revealing private information has not been as widely studied. In this context, two major types of approaches are considered: homomorphic encryption (HE)-based approaches and multi-party computation (MPC)-based approaches. These methods offer distinct advantages and are explored to address the privacy concerns associated with traditional regression learning algorithms.

\paragraph{HE-Based Approaches.}
Graepel et al.~\cite{graepel2012ml} used Somewhat Homomorphic Encryption that supports only a fixed number of multiplications to solve a linear system in a pricacy-preserving manner via gradient descent.
Ogilvie et al.~\cite{ogilvie2020improved} achieved privacy-preserving ridge regression training from homomorphic encryption via the Simiplified Fixed Hessian method.

We develop a new algorithm for non-linear regression tasks, the improved version of which can be computed by the same process of linear regression over transformalied data.

\paragraph{MPC-Based Approaches.} Nikolaenko et al.~\cite{nikolaenko2013privacy} presented an MPC-based protocol for privacy-preserving linear regression training, which combined a linear homomorphic encryption with the garbled circuit construction. Hall et al.~\cite{hall2011secure} also proposed an MPC-based protocol for training linear regression model, based on secret sharing under the semi-honest model.

However, the MPC-based approaches incur large communication overhead and require at most one of these two parties is malicious.

\subsection{Contributions}
Our main contribution is to present a new algorithm of privacy-preserving regression training using simplified fixed Hessian minimisation. Our starting point is based on a simple 2-layer neural newtork shared the same architecture as logistic regression. We derived a simplified fixed Hessian for our new algorithm and improve it to be free from computing the sigmoid function, which share the same calculation circul as linear regression. The resulting improved version can model non-linear relations between valuables.

We implement our new algorithm for regression tasks on both a small-scale dataset and a large-scale one in $\texttt{HEAAN}$.

As an additional contribution, we observe that normalizing not only the sample but also the predictions has many advantages when implementing regression algorithms in the encrypted domain, and thus we strongly suggenst doing so. Although this is a small advice, to the best of our knowledge, this is the first time it has been mentioned in the privacy-preserving manner. Doing so can help linear regression to get rid of overfitting and eliminate the need for selecting the regularization parameter in ridge regression, which is challenging to implement in the encrypted state.

\section{Preliminaries}


\subsection{Fully Homomorphic Encryption}
Homomorphic encryption is a family of encryption schemes that allows one to evaluate an arithmetic circuit on ciphertexts without decryption. One particular usefull type of such schemes is the Fully Homomorphic Encryption (FHE) that supports functions of a unlimited multiplicative depth, namely capable to compute any function on encrypted data.  FHE schemes used to have a significant disadvantage of managing the magnitude of plaintext when attempt to applying FHE in real-world machine learning tasks. Fortunately, Cheon et al.~\cite{cheon2017homomorphic} proposed an efficient approach called $\texttt{CKKS}$ with  a $\texttt{rescaling}$ procedure, which is suitable for handling practical data types such as real numbers. The $\texttt{CKKS}$ scheme supports approximate arithmetic on floating point numbers, interpreting noise as part of the usual errors observed in the clear when perform arithmetic operations in floating point representation ( fractional ). $\texttt{CKKS}$ has a native encoding function arising from the canonical embedding, and like several other homomorphic encryption schemes, enables the encoding of multiple messages into a single plaintext (aka $\texttt{SIMD}$) , which supports the "slotwise" operations on the messages using the automapgraph. The open-source library $\texttt{HEAAN}$ implementing the $\texttt{CKKS}$ scheme presented by Cheon et al.~\cite{cheon2017homomorphic, cheon2018bootstrapping} is comprised of all the algorihtms to perform modern homomorphic operations. 
We denote by $\texttt{ct.m}$ a ciphertext that encrypts the message m. For additional details, including other, we refer the readers to ~\cite{cheon2017homomorphic, cheon2018bootstrapping}.

When applying the $\texttt{SIMD}$ parallel technique, many refined algorithms usefull in calculating complex things such as the gradient have been proposed.


\subsection{Simplified Fixed Hessian}
Newton-Raphson method (or Newton's Method) used primarily for finding the roots of a real-valued function has the disadvantage of expensive computation due to the need to calculate and invert the Hessian matrix. The Fixed Hessian Method developed by B\"ohning and Lindsay~\cite{bohning1988monotonicity} is a variant of Newton's Method that reduces the computational burden by using a fixed (constant) approximation of the Hessian matrix instead of recalculating it at every iteration. In 2018, Bonte and Vercauteren~\citep{IDASH2018bonte} proposed the Simplified Fixed Hessian (SFH) method for logistic regression training over the BFV homomorphic encryption scheme, further simplifieding the Fixed Hessian method  by constructing a diagonal Hessian matrix using the Gershgorin circle theorem. 

Both Fixed Hessian method and Simplified Fixed Hessian method would fail to work because of the singular Hessian substitute if the datasets were some high-dimensional sparse martix, such as the MNIST datasets. Moreover, the construction of simplified fixed Hessian require all elements in the datasets to be non-positive. If the datasets were normalized into the range $[-1, +1]$, the SFH method might probably crack break down since it didn't satisfy the convergence condition of the Fixed Hessian method. These two limitations were removed by the work of Chiang~\cite{chiang2022privacy} who proposed a faster gradient variant called $\texttt{quadratic gradient}$, generalizing the SFH to be invertible in any case.

Also, both works~\cite{bohning1988monotonicity, IDASH2018bonte} didn't give a systemtic way to find the  (Simplified) Fixed Hessian substitute. Indeed, most optimastic functions perhaps don't have such a $good$ constant bound for Hessian matrix. Chiang~\cite{chiang2022quadratic} presents one method to probe if there is a SFH for the objective function, which is a necessary condition. This means that the function to optimise could probabily still have a SFH but Chiang's method failed to find it out.

For any function $F(\mathbf x)$  twice differentiable, if the Hessian matrix $H$ with its element $ \bar h_{ki} $ is positive definiteness for minimization problems (or negative definiteness for maximization problems), SFH method need to first find a constant matrix $\bar{H}$ that satisfy $\bar{H} \le H$ in the Loewner ordering for a maximization problem or $H \le \bar{H}$ otherwise. Next, SFH can be obtained in the following form:
 \begin{equation*}
  \begin{aligned}
 B =
\left[ \begin{array}{cccc}
 \sum_{i=0}^{d} \bar h_{0i}    & 0  &  \ldots  & 0  \\
 0  &   \sum_{i=0}^{d} \bar h_{1i}  &  \ldots  & 0  \\
 \vdots  & \vdots                & \ddots  & \vdots  \\
 0  &  0  &  \ldots  &  \sum_{i=0}^{d} \bar h_{di}  \\
 \end{array}
 \right].
   \end{aligned}
\end{equation*}
We recall that SFH need the dataset to have positive entries and that it can still be not invertable.

Chiang's quadratic gradient~\cite{chiang2022privacy} also need to build a  diagnoal matrix $\tilde B$ ahead, but which could be not constant, from the Hessian matrix $H$ with entries $H_{0i}$ :
 \begin{equation*}
  \begin{aligned}
   \tilde B = 
\left[ \begin{array}{ccc}
  - \epsilon - \sum_{i=0}^{d} | H_{0i} |    & 0  &  \ldots   \\
 0  &   - \epsilon - \sum_{i=0}^{d} | H_{1i} |  &  \ldots    \\
 \vdots  & \vdots                & \ddots       \\
 0  &  0  &  \ldots    \\
 \end{array}
 \right], 
   \end{aligned}
\end{equation*}
where $\epsilon$ is a small constant (usually $1e - 8$) added to the denominator to ensure numerical stability and to prevent division by zero. We remark that $\tilde B$ meets the convergence condition of fixed Hessian method no matter the dataset has negative entries or not.




\subsection{Linear Regression Model}
We briefly review linear regression and all the discussion below can be found in most classic machine learning textbooks (e.g.,~\cite{murphy2012machine}).

Given a dataset of $n$ samples $x_i \in \mathbb{R}^{d}$ with $d$ features, and $n$ output variables $y_i \in \mathbb{R}$, the goal of linear regression is to find a weight vector $\boldsymbol{\beta} = [\beta_0, \beta_1, \cdots, \beta_d] \in \mathbb{R}^{(1 + d)}$ such that for each $i$ $$ y_i \approx \beta_0 + \beta_1x_{i1} + \cdots + \beta_dx_{id} ,$$ which can be transformed into an optimization problem to minimize~\footnote{when applying first-order gradient descent methods, the common practise is to averge the squared residuals: $L_0 = \frac{1}{n} \sum_{i = 1}^{n} (\boldsymbol{\beta}^{\top}\mathbf{x}_i - y_i)^2$}:
$$L_0 = \sum_{i = 1}^{n} (\boldsymbol{\beta}^{\top}\mathbf{x}_i - y_i)^2,$$ 
where $\mathbf{x}_i = [1 \ x_i]$ is the feature vector $x_i$ with a $1$ inserted at front by appending a 1 to each of the inputs.  Linear regression aims to find the best-fit line (or hyperplane in higher dimensions) that minimizes the sum of the squared differences between the observed values and the predicted values.

Unlike logistic regression, linear regression has a closed-form expression, given by: $\boldsymbol{\beta} = X^+\mathbf{y}$, where 
\begin{equation*}
  \begin{aligned}
  X =   \left[ \begin{array}{c}
 \mathbf{x}_{0}  \\
 \mathbf{x}_{1}  \\
 \vdots   \\
 \mathbf{x}_{d}  \\
 \end{array}
 \right]
  =
\left[ \begin{array}{cccc}
 {x}_{10}  &   {x}_{11}  &  \ldots  &  {x}_{1d}  \\
 {x}_{20}  &   {x}_{21}  &  \ldots  &  {x}_{2d}  \\
 \vdots         &   \vdots         &  \ddots  &  \vdots         \\
 {x}_{n0}  &  {x}_{n1}   &  \ldots  &  {x}_{nd}  \\
 \end{array}
 \right], 
  \
  \mathbf y =   \left[ \begin{array}{c}
 y_{0}  \\
 y_{1}  \\
 \vdots   \\
 y_{d}  \\
 \end{array}
 \right],
  \end{aligned}
\end{equation*}
 and $X^+$ denotes the Moore-Penrose pseudo-inverse of the dataset $X$.
The Moore-Penrose pseudo-inverse of a matrix $X$, denoted $X^{+}$, is a generalization of the inverse matrix and provides a way to obtain the least-squares solution, which minimizes the Euclidean norm of the residuals $||X^{\top}\boldsymbol{\beta} - \boldsymbol y||.$
The pseudo-inverse $X^+$ can be computed through singular value decomposition. However, this technique is difficult to achieve in the FHE manner. When the closed-form solution is computationally expensive or infeasible, especially for large datasets, iterative optimization algorithms such as gradient descent or Newton's method can be used instead, minimising the cost function $L_0$ directly.
The gradient and Hessian of the cost function $L_0(\boldsymbol{\beta})$ are given by, respectively:
\begin{equation*}
  \begin{aligned}
\nabla_{\boldsymbol{\beta}} L_0(\boldsymbol{\beta}) &= 2 \sum_i (\boldsymbol{\beta}^{\top}\mathbf{x}_i - y_i) \mathbf x_i, \\
\nabla_{\boldsymbol{\beta}}^2 L_0(\boldsymbol{\beta}) &= 2 \sum_i  \mathbf x_i^{\top} \mathbf x_i
= 2 X^{\top}X ,
 \end{aligned}  
\end{equation*}
where $X$ is the dataset. 
 
\subsubsection{Ridge Regression} 
Ridge regression is a type of linear regression that includes an $L2$ regularization term to penalize large coefficients, which helps prevent overfitting by shrinking the coefficients towards zero. The cost function $L_1$ for ridge regression is to minimize the sum of squared residuals with a penalty on the size of the coefficients~\footnote{when applying first-order gradient descent methods, the common practise is to averge the squared residuals: $L_1 = \frac{1}{n} \sum_{i = 1}^{n} (\boldsymbol{\beta}^{\top}\mathbf{x}_i - y_i)^2 + \frac{\lambda}{2n} \sum_{i = 0}^{d} \beta_i^2,$}
$$L_1 = \frac{1}{2}\sum_{i = 1}^{n} (\boldsymbol{\beta}^{\top}\mathbf{x}_i - y_i)^2 + \frac{1}{2}\lambda \sum_{i = 0}^{d} \beta_i^2,$$ where $\lambda \ge 0$ is the regularization parameter, reflecting the extent to penalise large coefficients. The choice of the regularization parameter $\lambda$ is crucial and often requires cross-validation to determine.  

Like linear regression, ridge regression also has a closed form solution, given by: .
The gradient $\nabla_{\boldsymbol{\beta}} L_1(\boldsymbol{\beta}) $ and Hessian $\nabla_{\boldsymbol{\beta}}^2 L_1(\boldsymbol{\beta})$ of the cost function $L_1(\boldsymbol{\beta})$ are given by, respectively:
\begin{equation*}
  \begin{aligned}
\nabla_{\boldsymbol{\beta}} L_1(\boldsymbol{\beta}) &= \lambda  \boldsymbol{\beta} + \sum_i (\boldsymbol{\beta}^{\top}\mathbf{x}_i - y_i) \mathbf x_i, \\
\nabla_{\boldsymbol{\beta}}^2 L_1(\boldsymbol{\beta}) &= \lambda\boldsymbol{I} + \sum_i  \mathbf x_i^{\top} \mathbf x_i
=  \lambda\boldsymbol{I}  + X^{\top}X ,
 \end{aligned}  
\end{equation*}
where $\boldsymbol{I}$ is the identity matrix.

We highly suggest also normalizing the regressand into the range $[0, 1]$ or $[-1, +1]$. (helpful in overfitting)(easy to refresh ciphertext due to same coeffi range) 

We perfer privacy-preserving linear regression with normalized predictions over homomorphic ridge regression taining since it is difficult to select the regrulaer parameter $\lambda$ via CV in the encrypted domain. Linear regression with normalizing the predictions could reach the same goal of ridge regression and would not need to choose the regualr parameter. Therefore, we recommand the latter and will not further discuss ridge regression.
\section{Technical Details}
In this section, we first give our SFH for linear regression, then present our new algorithm for regression, and finally descript its improved version.

\subsection{Our Linear Regression Algorithm}
\label{ alg:our LR algorithm }
Ogilvie et al.~\cite{ogilvie2020improved} already give a SFH version for ridge regression, namely a diagonal matrix $\tilde{H} \in R^{}$ with diagonal elements given by
$$ \tilde{H}_{kk} = \lambda \mathbbm 1_{k \neq 0} + \sum_{j = 0}^d \sum_{i = 1}^n x_{ij}x_{ik}, $$
which draws heavily from the results of B\"ohning and Lindsay~\cite{bohning1988monotonicity} and Bonte and Vercauteren~\citep{IDASH2018bonte}. They assume all entries $x_{ij}$ are scaled to the range $[0, 1]$ in order to apply the Lemma in~\cite{IDASH2018bonte}. We point out here that the SFH $\tilde{H}$ could still be singular if the entries were scaled into the range $[-1, +1]$, happened sometimes in the real-world applications, and that their linear regression version $ \tilde{H}_{kk} = \sum_{j = 0}^d \sum_{i = 1}^n x_{ij}x_{ik}$ could be singular too even if all entries $x_{ij}$ were scaled to the range $[0, 1]$. Following the work of Chiang~\cite{chiang2022privacy, chiang2022multinomial}, we suggest using instead another form of SFH with the diagonal elements:  
$$ \tilde{B}_{kk} = | \lambda \mathbbm 1_{k \neq 0} + \sum_{j = 0}^d \sum_{i = 1}^n x_{ij}x_{ik} | + \epsilon. $$

We observe that normalizing not only the data but also the regressand would help to avoid overfitting and thus be HE-friendly to bootstrapping operation, reducing the burden of setting the regularisation parameter via cross validation.
Therefore, in this work we only discuss linear regression and explore its second-order Hessian-Based solution.

The Hessian for cost function $L_0$ is already ``fixed'' but still can be singular, which can be corrected by applying quadratic gradient. Following the construction method in~\cite{chiang2022quadratic}, from directly the Hessian $H$ we can obtain a diagonal matrix $\bar{B}$ with diagonal elements: $$\bar{B}_{kk} = \epsilon + 2 \sum_{j = 0}^d \sum_{i = 1}^n |x_{ij}x_{ik}|,$$ which has its inverse $\bar{B}^{-1}$ in any case. The inverse $\bar{B}^{-1}$ is also a diagonal matrix with diagonal elements $\frac{1}{ \bar{B}_{kk} } .$ 

Replacing the full Hessian $H$ for $L_0$ with our diagonal matrix $\bar B$, updates now becomes:
$$\boldsymbol{\beta} \leftarrow \boldsymbol{\beta} - \bar B^{-1} \nabla_{\boldsymbol{\beta}} L_0(\boldsymbol{\beta}) .$$

\subsection{Logistic Function for Regression}
Linear regression is based on the premise that the function  mapping the input variables (features) to the output variable (target) is linear and works very well in this situations. However, in real-world scenarios, the relationship is usually more complex. This complexity can arise due to non-linear interactions between variables, higher-order effects, and other intricate dependencies that a simple linear model might not capture. To address these more complex relationships, we propose a novel algorithm called LFFR (shorten for Logistic Function for Regression) with logistic regression-like struction, namely a 2-layer neural networks without any hidden layers using only the sigmoid activation function, to better model the data and improve prediction accuracy. 

\subsubsection{A Simple Assumption}  
We assume that, in some special regression cases, the output predict $y_i$ is a single likelihood between $0$ and $1$, which surely can be solved by a 2-layer nerual network with one singel neruo using simgoid as activation function. Seen as an extension of linear regression, our algorithm aims to minimize the following cost function $L2$, similar to $L_0$: 
$$L_2 = \sum_{i = 1}^{n} (\sigma( \boldsymbol{\beta}^{\top}\mathbf{x}_i ) - y_i)^2,$$
which is the same to a recently-proposed cost function but for classifications, $\text{square likilihood error}$ (SLE)~\cite{chiang2023privacyCNN, chiang2023privacy3layerNN}. Actually, $L_2$ is the exactly the Mean Squared Error (MSE) loss function, indicating the close relationship between MSE and SLE. 

The function $L_2$ has its gradient $g_2$ and Hessian $H_2$, given by:
\begin{equation*}
  \begin{aligned}
\nabla_{\boldsymbol{\beta}} L_2(\boldsymbol{\beta}) &= [ \sum_i 2(\sigma(i) - y_i )\sigma(i)(1 - \sigma(i))x_{i0}, \cdots, \sum_i 2(\sigma(i) - y_i )\sigma(i)(1 - \sigma(i))x_{id}]    \\
&= 
\sum_i 2(\sigma(i) - y_i )\sigma(i)(1 - \sigma(i))\mathbf x_i,    \\
\nabla_{\boldsymbol{\beta}}^2 L_2(\boldsymbol{\beta}) &= \sum_i \mathbf x_i^{\top} \Delta_i \mathbf x_i \\
&= X^{\top}SX , 
 \end{aligned}  
\end{equation*}
where  $\sigma(i) = \sigma(\boldsymbol{\beta}^{\top} \mathbf x_i),$ $\Delta_i = (4\sigma(i) - 6\sigma(i)^2 - 2y_i + 4y_i\sigma(i))\sigma(i)(1 - \sigma(i)),$ and $S$ is a diagonal matrix with diagonal entries $\Delta_i$.

According to~\cite{chiang2022privacy}, to find a simplified fixed Hessian for $H_2$, we should try to find a good uppder constant bound for each $\Delta_i.$ Since $0 \le \sigma(i) \le 1$,  $0 \le y_i \le 1$ and $\Delta_i$ is a polynomial of $\sigma(i)$ and $y_i$, $\Delta_i$ surely has its point to reach the maximium. Maximising $\Delta_i$ is a problem of unconstrained optimization. To do that, we need to calculate the first-order partial derivatives of $\Delta_i$ and have them equal to $0$. Solving the resulting equations, we obtain the unique solution to maximise $\Delta_i$:
$$\sigma(i) = 0.5, y_i = 0.5, \Delta_i = 0.125 = 1/8 .$$

Moreover, it is necessary to evaluate the function values at the boundary points, and finally, we obtain the final maximum value $\Delta_i \le 0.155$ when $\sigma(i) = 0.386$ and $y_i = 0$.

So far, we compute our simplified fixed Hessian $\bar B$ for $L_2.$ That is, a diagonal matrix with diagonal elements $\bar B_{kk}$ shown as:
$$\bar B_{kk} = \epsilon + 0.155\sum_{j = 0}^d \sum_{i = 1}^n |x_{ij}x_{ik}|.$$
 Our updates for LFFR are now given component wise by:
\begin{align}
\beta_j \leftarrow \beta_j + \frac{1}{\bar B_{jj}} \nabla_{\boldsymbol{\beta}} L_2(\boldsymbol{\beta}).
\label{ eq:LFFR }
\end{align}

\subsubsection{Real World Application}  
On the other hand, however, most real-world applications for regressions do not involve predicting probabilities, which instead often require forecasting continuous values such as house prices. In order to apply our LFFR algorithm to these tasks, we employ the thought of transformation from mathematics to transform the regression tasks predicting continuous values into regression problems predicting probabilities.

By leveraging mathematical transformation techniques, we convert the continuous output space into a probabilistic one, allowing our algorithm to effectively handle the original regression tasks within this new framework. This approach not only broadens the applicability of our LFFR algorithm but also takes advantage of the inherent properties of probability predictions, such as bounded outputs and well-defined uncertainty measures, enhancing the robustness and interpretability of the regression models in various real-world scenarios.

Moreover, we observe that normalizing the predicted values in regression tasks can help prevent overfitting to some extent and ensure that model weight coefficients remain within a small range. In privacy-preserving machine learning applications using homomorphic encryption, this normalization is beneficial to parameter settings for ciphertext refreshing. As a result, we strongly recommend normalizing predicted values even in linear regression models to maintain the efficacy and security of the encryption process.
To handle the regression tasks effectively, we apply a mathematical transformation to convert the continuous regression problem into a probabilistic prediction task. This transformation maps the continuous output space into a probability space.

Given the input features $X$ and the target values $\mathbf{y}$, our revised LFFR algorithm incorporating the normalization for prediction values consisst of the following $3$ steps:

\indent $\texttt{ Step 1:}$ 
We preprocess the dataset to ensure $X$ and $\mathbf y$ both in the required format. If the regression task were to predict continuous values other than likilihood, we should normalize the target values to fall within the $[0, 1]$ range. For example, we can iterate through the predicted values  $mathbb{y}$ to find the minimum $(y_{min})$ and maximum $(y_{max})$ values and scale each prediction $y_i$ to a common range $[0, 1]$ by the following formula: $$ \bar y_i = \frac{y_i - y_{min}}{y_{max} - y_{min} + \epsilon}, $$ where $\bar y_i$ is the normalized value of $y_i$.
The training data $X$ should also be scaled to a fixed range, usually $[0, 1]$ or $[-1, 1]$. 

\indent $\texttt{ Step 2:}$ 
Using the data $X$ and the new predictions $\bar{\mathbf{y}}$ consissting of normalized values $\bar y_i$, we train the LFFR model to fit the transformed dataset, updating the model parameter vector $\boldsymbol{\beta}$ by the formula~$($\ref{ eq:LFFR }$)$. After the completion of the training process, the trained model weights $\boldsymbol{\beta}_{trained}$ are obtained.

\indent $\texttt{ Step 3:}$ 
When a new sample $\mathbf{x}$  arrives, we first apply the same normalization technique used on the training data to $\mathbf{x}$, resulting in a normalized sample $\bar{ \mathbf{x} }.$ Then, $\boldsymbol{\beta}_{trained}$ is employed to compute the new sample's probability $prob = \sigma(\boldsymbol{\beta}_{trained}^{\top} \bar{\mathbf{x} })$. Finally, we utilize the inverse function of the normalization to map the likelihood $prob$ back to continuous values, obtaining the prediction $y_i$ in the original continuous space using $y_{min}$ and $y_{max}:$
$$y_i = (y_{max} - y_{min} + \epsilon) \cdot prob +  y_{min}.$$

\subsubsection{Limitation and Soloution}
One main limitation of our LFFR algorithm is that the predicted values for new, previously unseen data will not exceed the range $[y_{min}, y_{max}]$ of the predicted values in the training set. There are two possible solutions to addressing this limitation:
\begin{enumerate}
    \item one simple solution is to scale the predicted values in the training set to a smaller interval within the  $[0, 1]$ range, such as $[0.25, 0.75],$ which would effectively enlarges the range of predicted values for the new sample arrived.

    \item the other is to dynamically adjust the scaling parameters for the predicted values of new data, similar to the norm layer in deep learning where $\bar y = ay + b$, with $a$ and $b$ both being trainable parameters.
\end{enumerate}

Our LFFR algorithm has another significant drawback: the increased computational load due to the frequent use of the sigmoid function. This issue is particularly pronounced in privacy-preserving computing environments, where additional computation can lead to increased latency and resource consumption, further complicating the implementation and efficiency of the algorithm.

\subsection{The Improved LFFR Version}
After converting the contiunous predictions $\mathbf{y}$ into likilihoods $\bar{ \mathbf{y} }$, our LFFR algorihtm aims to solve a system of non-linear equaltions:
\begin{align}
\begin{cases}
   \sigma(\boldsymbol{\beta}^{\top}\mathbf{x}_1) \approx \bar{y}_1 = \frac{y_1 - y_{\min}}{y_{\max} - y_{\min} + \epsilon}, \\
   \sigma(\boldsymbol{\beta}^{\top}\mathbf{x}_2) \approx \bar{y}_2 = \frac{y_2 - y_{\min}}{y_{\max} - y_{\min} + \epsilon}, \\
   \hspace{3cm} \vdots \\
   \sigma(\boldsymbol{\beta}^{\top}\mathbf{x}_n) \approx \bar{y}_n = \frac{y_n - y_{\min}}{y_{\max} - y_{\min} + \epsilon}.
\end{cases}
\label{ eq:non-linear system }
\end{align}

It is important to point out that the converting function to normilize predictions could just be a linear bijective function rather than otherwise. For instance, if we use the simgoid function to map $y_i$ into the range $[0, 1]$, our LFFR algorihtm will reduce to the original linear regression since $\sigma(\boldsymbol{\beta}^{\top}\mathbf{x}_i) \approx \sigma(y_i) \Rightarrow \boldsymbol{\beta}^{\top}\mathbf{x}_i \approx y_i$ is the formalus of linear regression.

The system of equations~$($\ref{ eq:non-linear system }$)$ can be transformed into:
\begin{align}
\begin{cases}
   \sigma^{-1}(\sigma(\boldsymbol{\beta}^{\top}\mathbf{x}_1)) = \boldsymbol{\beta}^{\top}\mathbf{x}_1 \approx \sigma^{-1}(\bar{y}_1) = \sigma^{-1}( \frac{y_1 - y_{\min}}{y_{\max} - y_{\min} + \epsilon}), \\
   \sigma^{-1}(\sigma(\boldsymbol{\beta}^{\top}\mathbf{x}_2)) = \boldsymbol{\beta}^{\top}\mathbf{x}_2 \approx \sigma^{-1}(\bar{y}_2) = \sigma^{-1}( \frac{y_2 - y_{\min}}{y_{\max} - y_{\min} + \epsilon}), \\
   \hspace{3cm} \vdots \\
   \sigma^{-1}(\sigma(\boldsymbol{\beta}^{\top}\mathbf{x}_n)) = \boldsymbol{\beta}^{\top}\mathbf{x}_n \approx \sigma^{-1}(\bar{y}_n) = \sigma^{-1}( \frac{y_n - y_{\min}}{y_{\max} - y_{\min} + \epsilon}),
\end{cases}
\label{ eq:non-linear system transformed }
\end{align}
where $\sigma^{-1}(y) = \ln(\frac{y}{1-y}) $ for $0 \le y \le 1$ is the inverse of the sigmoid function, known as the logit function. 

The system of equations~$($\ref{ eq:non-linear system transformed }$)$ can be seen as linear regerssion over a new dataset consissting of the original samples and the new likilhood predcitions converted from the original ones $y_i$ by $\sigma^{-1}( \frac{y_i - y_{\min}}{y_{\max} - y_{\min} + \epsilon}) ,$ which can be solved via our linear regression algorithm in section . Since $\sigma^{-1}(0)$ and $\sigma^{-1}(1)$ are negative infinity and positive infinity, respectively, for numical stabdity we cannot normalize original predicted values into the range containing $0$ and $1$. Therefore, we introduce a parameter 
$\gamma$ to control the range of the newly generated predictions within the interval $[0, 1]$, excluding $0$ and $1$, where $0 < \gamma < 1.$ The new range for probability predictions will be $[0.5 - \gamma/2, 0.5 + \gamma/2] .$

Our improved LFFR algorithm, which removes the limitations of the original LFFR and is described in Algorithm~\ref{ alg:our improved LFFR algorithm }, consists of the following $3$ steps:

\indent $\texttt{ Step 1:}$ 
Whether or not the task is to predict likilihood, we normalize the prediction $y_i$ first and transform it to a new one $\bar y_i$, with the parameter $\gamma$ by the mapping: $\bar y_i = sigma^{-1}( \frac{y_i - y_{\min}}{y_{\max} - y_{\min} + \epsilon}*\gamma + 0.5 - \gamma/2) .$  
We assume that the dataset $X$ has already been normalized into a certain range like $[-1, +1].$
The normalized data $x_i$ with the new transformed predictions $\bar{y}_i$ is composed of a new regression task to predict some simulation likilihood, which can be solved by our linear regression algorithm in Section~\ref{ alg:our LR algorithm }. 

\indent $\texttt{ Step 2:}$ 
We apply our linear regression algorithm to address the new regression task of predicting simulation likelihoods $\bar{y}_i$, obtaining a well-trained weight vector $\boldsymbol{\beta}$. This step is the same as that of our LFFR algorithm.

\indent $\texttt{ Step 3:}$ 
For some new, already normalized sample $\mathbf{x}$ to predict, we first use the well-trained weight vector $\boldsymbol{\beta}$ to compute its simulatedtion likilihood $prob = \sigma(\boldsymbol{\beta}^{\top} \bar{\mathbf{x} }).$ Note that $prob \in [0, 1]$ could exceed the range $[0.5 - \gamma/2, 0.5 + \gamma/2]$ and therefore let our improved LFFR algorihtm to predict continuous values outside the original prediction range $[y_{min}, y_{max}]$.

Then, we utilize the inverse function of the new  normalizetion mapping to convert the likelihood $prob$ back to the true prediction value $\bar y$ in the original continuous space using $y_{min}$ and $y_{max}$:
$$\bar y = (y_{max} - y_{min} + \epsilon) \cdot (prob - 0.5 + \gamma/2)/\gamma +  y_{min}.$$

\begin{algorithm}[htbp]
    \caption{Our Improved LFFR algorithm }
    \label{ alg:our improved LFFR algorithm  }
     \begin{algorithmic}[1]
        \Require Normalized dataset $ X \in \mathbb{R} ^{n \times (1+d)} $, target values $ Y \in \mathbb{R} ^{n \times 1} $, float number parameter $ \gamma \in \mathbb{R} (\texttt{set to 0.5 in this work}) $, and the number  $\kappa$ of iterations
        \Ensure the weight vector $ \boldsymbol{\beta} \in \mathbb{R} ^{(1+d)} $ 

        \State Set $ \boldsymbol{\beta} \gets \boldsymbol 0$
        \Comment{$\boldsymbol{\beta} \in \mathbb{R}^{(1+d)}$}        
        
        \State Set $\bar B \gets \boldsymbol 0$
        \State Set $\bar H \gets 2X^{\top}X$
        \Comment{$\bar H \in \mathbb{R}^{(1+d) \times (1+d)}$}
                
           \For{$i := 0$ to $d$}
              \State $\bar B[i][i] \gets \epsilon$
              \Comment{$\epsilon$ is a small positive constant such as $1e-8$}
              \For{$j := 0$ to $d$}
                 \State $ \bar B[i][i] \gets \bar B[i][i] + |\bar H[i][j]| $
              \EndFor
              \State $ \bar B[i][i] \gets 1 / \bar B[i][i] $
           \EndFor
       
        \State Set $ \bar Y \gets \boldsymbol 0$
        \Comment{$\boldsymbol{\beta} \in \mathbb{R}^{(1+d)}$} 
        \State Set $ Y_{\min} \gets \min(Y) $
        \State Set $ Y_{\max} \gets \max(Y) $
        \For{$i := 0$ to $d$}
           \State $ \bar Y[i] \gets  \sigma^{-1}( \frac{Y[i] - Y_{\min}}{Y_{\max} - Y_{\min} + \epsilon}*\gamma + 0.5 - \gamma/2) $
        \EndFor                 
                 
        \For{$k := 1$ to $\kappa$}
           \State Set $Z \gets \boldsymbol 0 $
           \Comment{$Z \in \mathbb{R}^{n}$  will store  the dot products     }
           \For{$i := 1$ to $n$}
              \For{$j := 0$ to $d$}
                 \State $ Z[i] \gets  Z[i] +  \boldsymbol{\beta}[j] \times  X[i][j] $ 
              \EndFor
           \EndFor

           \State Set $\boldsymbol g \gets \boldsymbol 0$
           \For{$j := 0$ to $d$}
              \For{$i := 1$ to $n$}
                 \State $\boldsymbol g[j] \gets \boldsymbol g[j] + 2 \times (Z[i] - \bar Y[i]) \times X[i][j] $
              \EndFor
           \EndFor
           
           \For{$j := 0$ to $d$}
              \State $ \boldsymbol{\beta}[j] \gets \boldsymbol{\beta}[j]  -  \bar B[j][j] \times \boldsymbol g[j] $
           \EndFor
           
        \EndFor
      
        \State \Return $ \boldsymbol{\beta} $
        \Comment{For a new normalized sample $\mathbf x$, The algorithm output the prediction: $(\sigma(\boldsymbol{\beta}^{\top}\mathbf{x}) - 0.5 + \gamma/2)/\gamma(Y_{\max} - Y_{\min} + \epsilon) + Y_{\min}$ }
        \end{algorithmic}
\end{algorithm}

\subsection{Comparison in The Clear}

We assess the performance of four algorithms: our LFFR algorithm (denoted as  LFFR), our improved LFFR algorithm (ImprovedLFFR), our linear regression algorihtm (LR) and its variant with normalized predictions (YnormdLR), in the clear setting using the Python programming language. To evaluate the convergence speed,  we choose the loss function, mean squared error (MSE), in the training phase  as the only indicator.  Random datasets with different parameters are generated to evaluate these four algorithms: Let the data has $n$ samples $x_i$ normalized into the range $[-1, +1]$, with $d$ features; $d$ float-point numbers $r_i$ from the range $[-1, +1]$ are randomly generated to present the linear relationship between the sample and its prediction, with a noise $nois$ generated follows a Gaussian distribution  $\mathcal{N}(0, \sigma^2)$ with a mean of $0$ and a variance of $\sigma^2$. Therefore, the prediction $y_i$ for $x_i$ is $y_i = nois + \sum_{k = 1}^d r_k x_{ik}.$ Our linear regression variant will in advance normalize the predictions, record the minimum and maximum valuse of the prediction and use them in the inference phase to compute the real preditions in the original space. 

\textbf{Analysis and Discussion} 



Based on a recent theory~\cite{chiang2023activation} interpreting neural network models, our algorithm can approximate more complex nonlinear relationships. It can be viewed as a polynomial regression model or an advanced linear regression model over complex combinations of features:
$$
y_i = \sigma(\boldsymbol{\beta}^{\top} \bar{\mathbf{x} }) = poly(\boldsymbol{\beta}^{\top} \bar{\mathbf{x} })
,$$
where $poly(\cdot)$ is the polynomial approximation of the sigmoid function $\sigma(\cdot)$.

\section{Secure Regression }
When employed in the encrypted domain, our LFFR algorithm involving the calculation of the logistic function faces a difficult dilemma that no homomorphic schemes are capable of directly calculating the sigmoid function. A common solution is to replace the sigmoid function with a polynomial approximation. Several works~\cite{chiang2024simple, chiang2022polynomial} have been conducted to find such polynomials and current state-of-the-art technique~\cite{cheon2022homomorphicevaluation} can approximate the sigmoid function over a pretty large intervals with reasonable precious, such as over the range $[-3000, +3000].$ In this work our implementation, we just call a function named `` $\texttt{polyfit($\cdot$)}$ ''  in  the Python package Numpy, utilizing the least squares approach to approximate the sigmoid function over the interval $[-5, 5]$ by a degree 3 polynomial: 
$ g(x) = 0.5 + 0.19824 \cdot x  - 0.0044650 \cdot x^3  .$

The difficulty in applying the quadratic gradient  is to invert the diagonal matrix $ \tilde {B} $ in order to obtain  $ \bar {B} $. We leave the computation of  matrix $ \bar {B} $ to  data owner and let the data owner upload the ciphertext encrypting the $ \bar {B} $ to the cloud. Since  data owner has to prepare the dataset and normalize it, it would also be practicable for the data owner to calculate the $ \bar {B} $  owing to no leaking of sensitive data information.

\subsection{Database Encoding Method} \label{basic he operations} 

Given the training dataset $\text X \in \mathbb R^{n \times (1+d)}$ and  training label $\text Y \in \mathbb R^{n \times 1}$, we adopt the same method that Kim et~al.~\cite{kim2018logistic} used to encrypt the  data matrix consisting of the training data combined with training-label information into a single ciphertext $\text{ct}_Z$. The weight vector $\boldsymbol{\beta}^{(0)}$ consisting of zeros and the diagnoal elements of $\bar B$ are copied $n$ times to form two matrices. The data owner then encrypt the two matrices into two ciphertexts $\text{ct}_{\boldsymbol{\beta}}^{(0)}$ and $\text{ct}_{\bar B}$, respectively. The ciphertexts $\text{ct}_Z$, $\text{ct}_{\boldsymbol{\beta}}^{(0)}$ and $\text{ct}_{\bar B}$ are as follows:
\begin{equation*}
 \begin{aligned}
\text{ct}_X &= Enc
\left[ \begin{array}{cccc}
 1  &   x_{11}  &  \ldots  &  x_{1d}  \\
 1  &   x_{21}  &  \ldots  &  x_{2d}  \\
 \vdots         &   \vdots         &  \ddots  &  \vdots         \\
 1  &  x_{n1}   &  \ldots  &  x_{nd}  \\
 \end{array}
 \right], 
&\text{ct}_Y = Enc
\left[ \begin{array}{cccc}
 y_{1}  &   y_{1}  &  \ldots  &  y_{1}  \\
 y_{2}  &   y_{2}  &  \ldots  &  y_{2}  \\
 \vdots         &   \vdots         &  \ddots  &  \vdots         \\
 y_{n}  &   y_{n}  &  \ldots  &  y_{n}  \\
 \end{array}
 \right],   \\
 \text{ct}_{\boldsymbol{\beta}}^{(0)} &= Enc
\left[ \begin{array}{cccc}
 \beta_{0}^{(0)}  &   \beta_{1}^{(0)} &  \ldots  &  \beta_{d}^{(0)} \\
 \beta_{0}^{(0)}  &   \beta_{1}^{(0)} &  \ldots  &  \beta_{d}^{(0)}  \\
 \vdots         &   \vdots         &  \ddots  &  \vdots         \\
 \beta_{0}^{(0)}  &   \beta_{1}^{(0)} &  \ldots  &  \beta_{d}^{(0)}  \\
 \end{array}
 \right],  
& \text{ct}_{\bar B} = Enc
\left[ \begin{array}{cccc}
  \bar B_{[0][0]}    & \bar B_{[1][1]}  &  \ldots  & \bar B_{[d][d]}  \\
  \bar B_{[0][0]}    & \bar B_{[1][1]}  &  \ldots  & \bar B_{[d][d]}  \\
 \vdots  & \vdots                & \ddots  & \vdots     \\
  \bar B_{[0][0]}    & \bar B_{[1][1]}  &  \ldots  & \bar B_{[d][d]}  \\
 \end{array}
 \right], 
 \end{aligned}
\end{equation*}
where $\bar B_{[i][i]}$  is the diagonal element of  $\bar B$ that can be built from $\frac{1}{8}X^{\top}X$ for our LFFR algorithm and from $2X^{\top}X$ for our linear regression and Improved LFFR algorithms.

\subsection{The Usage Scenarios} 
Supposing two different roles including data oweners (e.g. hospital or individuals) and cloud computing service providers (e.g. Amazon, Google or Microsoft), our proposed algorithms can be employed to two main situations:
\begin{enumerate}
    \item In a private cloud machine learning, a single data owner uploads its private data after encryption with an HE scheme, and later downloads an encrypted model from the cloud server that has performed the machine learining training on the encrypted data. Homomorphic encryption allows the encrypted sensitive information to be stored and computed in a cloud server without the request for decryption keys, requiring no participation of data owners. By encrypting their data with public keys, data owners can directly upload the resulting ciphertexts to the cloud service and possess excludly the private key of the HE scheme. As a result, the cloud has only the public parameters and public keys and no private data is leaked.  

    \item Our algorithms can also be employed in the multiple data owner setting. In this application scenario, each data owner encrypts its data with the same designed HE scheme using the same shared public keys and uploads the resulting ciphertexts to the cloud, which performs the same machine learning training task as before. A well-trained model encrypted with the same HE system can be download from the cloud, decrypted by a group of entities that possesses their own share of the private key. These entities can jointly generate the public keys with their random shares of the private key using an additional key sharing protocol. 
\end{enumerate}

In either case, no private information is revealed as long as the underlying HE scheme is not broken and the secret key is not disclosed. Even if the cloud is compromised the data is till secure. Hence privacy-preserving machine learning training could be an ultimate solution for analyzing private data while keeping privacy.
\subsection{The Whole Pipeline} 
The full pipeline of privacy-preserving logistic regression training consists of the following steps:

\indent $\texttt{ Step 1:}$ 
The client prepares two data matrices, $\bar B$ and $Z$, using the training dataset.

\indent $\texttt{ Step 2:}$ 
The client then encrypts three matrices, $\bar B$, $Z$ and weight matrix $W^{(0)}$, into three ciphertexts $ \text{ct}_Z $, $ \text{ct}_{W}^{(0)} $ and $ \text{ct}_{\bar{B}} $, using the public key given by the third party under the assigned HE system.   We recommend adopting the zero matrix as the initial weight matrix, if a already-trained weight matrix is not provided.

\indent $\texttt{ Step 3:}$
The client finally uploads the ciphertexts to the cloud server and finished its part of precess, waiting for the result.

\indent $\texttt{ Step 4:}$
The public cloud begins its work by evaluating the gradient ciphertext $ \text{ct}_{g}^{(0)} $ with $ \text{ct}_Z $ and $ \text{ct}_{W}^{(0)} $ using various homomorphic operations. Kim et al.~\cite{kim2018logistic} give a full and detail description about homomorphic evaluation of the gradient descent method.

\indent $\texttt{ Step 5:}$
The public cloud computes the quadratic gradient with one homomorphic multiplication from $ \text{ct}_{\bar{B}} $ and $ \text{ct}_{g}^{(0)} $, resulting in the ciphertext $ \text{ct}_{G}^{(0)} $.

\indent $\texttt{ Step 6:}$
This step is to update the weight ciphertext with the quadratic gradient ciphertext using our enhanced mini-batch NAG method. This will consume some modulus level of the weight ciphertext, finally .

\indent $\texttt{ Step 7:}$
This step checks if the remaining modulus level of the weight ciphertext enable another round of updating the weight ciphertext. If not, the algorithm would bootstrap only the weight ciphertexts using some public keys,  obtaining a new ciphertext encrypting the same weight but with a large modulus. 

\indent $\texttt{ Step 8:}$
This public cloud completes the whole iterations of homomorphic LR training, obtain the resulting weight ciphertext, and finishes its work by returning a ciphertext encrypting the updated modeling vector to the client. 

Now, the client could decipher the received ciphertext using the secret key and has the LR model for its only use.


Han et al.~\cite{han2018efficient} give a detailed description of their HE-friendly LR algorithm with HE-optimized body of the iteration loop using HE programming in the encrypted domain. Our enhanced mini-batch NAG method is significantly similar to theirs except for needing one more ciphertext multiplication between the gradient ciphertext and the uploaded ciphertext encrypting $\bar B_i$ for each mini batch, and therefore please refer to~\cite{han2018efficient} for more information.


The pulbic cloud takes the three ciphertexts $\text{ct}_Z$, $\text{ct}_{\boldsymbol{\beta}}^{(0)}$ and $\text{ct}_{\bar B}$ and evaluates the enhanced NAG algorithm to find a decent weight vector by updating the vector $\text{ct}_{\boldsymbol{\beta}}^{(0)}$. Refer to~\cite{kim2018logistic} for a detailed description about how to calculate the gradient by HE programming.

\section{Experiments}
\paragraph{Implementation}  We employ our linear regression algorithm with normalized predictions, our LFFR and Improved LFFR algorithms to implement privacy-preserving regression training based on HE with the  library  $\texttt{HEAAN}$. The reason why we do not implement linear regression without normalizeing the predictions is the difficult to setting varying HE paramters for various regression tasks. The C++ source code is publicly available at \href{https://github.com/petitioner/HE.LFFR}{https://github.com/petitioner/HE.LFFR} . All the experiments on the ciphertexts were conducted on a public cloud with $1$ vCPUs and $50$ GB RAM.

\paragraph{Datasets}
In order to evaluate the effectiveness of our implementations, we use two classic datasets available through TensorFlow that already divides these them into training dataset and testing dataset: a small-scale dataset called Boston Housing~\footnote{This dataset may have an ethical problem.} consisting of $506$ records with $13$ covariates and a large-scale dataset named California Housing with $20,640$ samples with $8$ features each. Both datasets have a scalar target variable: the median values of the houses at certain location. Note that the final weights for the original linear regression algorihtm on the two datasets are so different from each other that it is difficult to refreshing the weight ciphertext for the large-scale California Housing dataset using the same parameter setting for Boston Housing dataset. We even doubt it might not be practial to set one common parameter for the original linear regression for any real-world tasks. As such, we strongly encourage to normalize the predictions of datasets into small ranges like $[-1, +1]$ or $[0, 1]$.

\paragraph{Parameters}
While the two datasets involved in our experiments have different sizes to each other, we can still set the same $\texttt{HEAAN}$ parameters for them: $logN = 16$, $logQ = 1200$, $logp = 30$, $slots = 32768$, which ensure the security level $\lambda = 128$.  Please refer to \cite{kim2018logistic} for details on these parameters. In such setting, the Boston Housing dataset is encrypted into a single ciphertext while the California Housing dataset into $33$ ciphertexts. Since our LFFR algorithm need to compute the sigmoid function in the encrypted state, it consumes more modulus and hence need to refresh its weight ciphertext every $ $  iterations while our Improved LFFR algorithm only every $ $ iterations. For our improved LFFR algorihtm, selecting the optimimum parameter $\gamma$ is beyond the scope of this work and we just set it to $0.5.$ Our LFFR algorihtm uses the degree $3$ polynomial $g(x)$ to approximate the sigmoid function.

\paragraph{Results}
TensorFlow ($v2.16.1$) already divide the two experimenting datasets into training dataset and testing dataset with the parameters $test\_split=0.2$ and $seed=113$. We evaluate our LFFR algorithm and its improved version on the two encrypted training datasets, using the decrypted weights to assess the performance on both the training and testing sets. 

\paragraph{Microbenchmarks}

\paragraph{Discussion}

Our algorithm would likely show improved performance in terms of convergence speed if we applied a quadratic gradient with an optimized learning rate configuration to the SFH.

\section{Conclusion}

In this paper, we developed an efficient algorithm named LFFR for homomorphic regression utilizing the logistic function. This algorithm can capture more complex relationships between input values and output predictions than traditional linear regression.

\bibliography{HE.LFFR}
\bibliographystyle{apalike}

\end{document}